\documentclass[letterpaper, 10 pt, conference]{ieeeconf}
\usepackage{epsfig}
\usepackage{psfrag,graphicx,color}
\usepackage{amsmath}
\usepackage{algorithm2e}
\usepackage[mathscr]{eucal}

\IEEEoverridecommandlockouts
\usepackage[normalem]{ulem}
\usepackage{amsmath,amsfonts}
\usepackage{epsfig,psfrag,graphicx,color}
\font\bfmath=cmmib10
\mathchardef\Gamma="7100
\mathchardef\Delta="7101
\mathchardef\Theta="7102
\mathchardef\Lambda="7103
\mathchardef\Xi="7104
\mathchardef\Pi="7105
\mathchardef\Sigma="7106
\mathchardef\Upsilon="7107
\mathchardef\Phi="7108
\mathchardef\Psi="7109
\mathchardef\Omega="710A

\mathchardef\alpha="710B
\mathchardef\beta="710C
\mathchardef\gamma="710D
\mathchardef\delta="710E
\mathchardef\epsilon="710F
\mathchardef\zeta="7110
\mathchardef\eta="7111
\mathchardef\theta="7112
\mathchardef\iota="7113
\mathchardef\kappa="7114
\mathchardef\lambda="7115
\mathchardef\mu="7116
\mathchardef\nu="7117
\mathchardef\xi="7118
\mathchardef\pi="7119
\mathchardef\rho="711A
\mathchardef\sigma="711B
\mathchardef\tau="711C
\mathchardef\upsilon="711D
\mathchardef\phi="711E
\mathchardef\chi="711F
\mathchardef\psi="7120
\mathchardef\omega="7121
\mathchardef\epsilon="7122

\mathchardef\varepsilon="7122
\mathchardef\vartheta="7123
\mathchardef\varpi="7124
\mathchardef\varrho="7125
\mathchardef\varsigma="7126
\mathchardef\varphi="7127
\mathchardef\imath="717B
\mathchardef\jmath="717C

\def\bfq{{\mbox{\boldmath $q$}}}

\def\bfsigma{{\mbox{\boldmath $\sigma$}}}

\def\smallbfW{{\raise1.5pt\hbox{\mbox{\boldmath $_W$}}}}

\let\ts=\thinspace

\def\mypsfrag#1#2#3#4#5{
        \begin{figure}[htp]
           \begin{center}
              {\leavevmode
                 {\includegraphics[width=#1truecm]{#2.eps}}
              }
           \end{center}
           \vspace{#3}
           \caption{#4}
           \vspace{-10pt}
           \label{#5}
        \end{figure}
}

\def\mypsfragtwocolumn#1#2#3#4#5{
        \begin{figure*}[htp]
           \begin{center}
              {\leavevmode
                 {\includegraphics[width=#1truecm]{#2.eps}}
              }
           \end{center}
           \vspace{#3}
           \caption{#4}
           \label{#5}
        \end{figure*}
}

\def\my4psfrag#1#2#3#4#5#6#7#8{
        \begin{figure}[htp]
        \begin{center}
	        \begin{tabular}[h]{c c}
              {\leavevmode{\includegraphics[width=#1truecm]{#2.eps}}}
              &
              {\leavevmode{\includegraphics[width=#1truecm]{#3.eps}}} \\
              {\leavevmode{\includegraphics[width=#1truecm]{#4.eps}}}
              &
              {\leavevmode{\includegraphics[width=#1truecm]{#5.eps}}}
   	     \end{tabular}
           \vspace{#6}
           \caption{#7}
           \label{#8}
        \end{center}
        \end{figure}
}

\def\mydouble4psfrag#1#2#3#4#5#6#7#8{
        \begin{figure*}[htp]
        \begin{center}
            \begin{tabular}[h]{c c}
              {\leavevmode{\includegraphics[width=#1truecm]{#2.eps}}}
              &
              {\leavevmode{\includegraphics[width=#1truecm]{#3.eps}}} \\
              {\leavevmode{\includegraphics[width=#1truecm]{#4.eps}}}
              &
              {\leavevmode{\includegraphics[width=#1truecm]{#5.eps}}}
         \end{tabular}
           \vspace{#6}
           \caption{#7}
           \label{#8}
        \end{center}
        \end{figure*}
}

\input{texdraw}
\definecolor{myorange}{rgb}{1,0.5,0}

\begin{document}

\title{\LARGE \bf Assistive robot operated via P300-based Brain Computer Interface}
\author{Filippo Arrichiello, Paolo Di Lillo, Daniele Di Vito, Gianluca Antonelli, Stefano Chiaverini%*
\thanks{Authors are with the Department of Electrical and Information Engineering of the University of Cassino and Southern Lazio,
		Via G. Di Biasio 43, 03043 Cassino (FR), Italy
		{\tt\small \{f.arrichiello, pa.dilillo, d.divito, antonelli, chiaverini\}@unicas.it}}
%\thanks{*Authors are in alphabetic order.}
}

%\markboth{A decentralized controller-observer scheme for multi-agent systems}%
%{Antonelli, Arrichiello, Caccavale, Marino}

\maketitle

\begin{abstract}
In this paper we present an architecture for the operation of an assistive robot finally aimed at allowing users with severe motion disabilities to perform manipulation tasks that may help in daily-life operations. The robotic system, based on a lightweight robot manipulator, receives high level commands from the user through a Brain-Computer Interface based on P300 paradigm. The motion of the manipulator is controlled relying on a closed loop inverse kinematic algorithm that simultaneously manages multiple set-based and equality-based tasks. The software architecture is developed relying on widely used frameworks to operate BCIs and robots (namely, BCI2000 for the operation of the BCI and ROS for the control of the manipulator) integrating control, perception and communication modules developed for the application at hand. Preliminary experiments have been conducted to show the potentialities of the developed architecture.

\end{abstract}

\section{Introduction} \label{sec:intro}
Research and development of robotic assistive technologies has gained tremendous momentum in the last decade, due to several factors such as the maturity level reached by several technologies, the advances in robotics and AI, and the fact that more than 700 million of persons have some kind of disability or handicap~\cite{united2007exclusion}. For many people with mobility impairments, essential and simple tasks, as dressing or feeding, require the assistance of dedicated people; thus, the use of devices providing independent mobility can have a large impact on their quality of life~\cite{hoenig2003does}.

In this perspective, different classes of robotic devices can be considered as lightweight robotic arms that may help in manipulation tasks, intelligent semi-autonomous powered wheelchairs to help in mobility tasks, or wheelchair-mounted robotic manipulator to help in both the issues. From the user perspective, the operation mode of such systems may depend on the level of autonomy provided by/required to the robotic system; from the device perspective, this correspond to different control modes that vary from shared to supervisory control. In shared mode, the user is involved in the control loop of the system by continuously generating high-frequency motion commands; such commands are then translated from the control software in low-level functions after applying all the safety policies. In supervisory mode, the user provides high level low-frequency commands (e.g., to start/stop actions) and the system operates in complete autonomy; the control software must generate motion directives that realize the required action while taking into account safety, comfort and efficiency. 

The operation modes of the robotic devices are strictly connected to the Human-Machine Interface (HMI) used to generate and communicate commands. Among the different HMIs,  Brain-Computer-Interfaces (BCIs) represent a relatively new technology that has recently attracted large attention in view of the fact that BCIs may be used in the absence of motion capability of the user~\cite{mcfarland2008brain} with applications in different areas of assistive technologies as motor recovery, entertainment, communication and control~\cite{millan2010combining,moritz2016new}. Indeed, BCIs have been recently proposed to drive wheelchairs~\cite{bi2013eeg,carlson2013brain}, to guide robots for telepresence~\cite{escolano2012telepresence,leeb2015towards}, to control exoskeletons~\cite{frisoli2012new} and mobile robots~\cite{riechmann2015using,gandhi2014eeg}.

Most BCIs rely on non-invasive electroencephalogram (EEG) signals, i.e. the electrical brain activity recorded from electrodes placed on the scalp. By processing such signals, the BCIs may allow the generation of commands that can be used for the communication with a software interface. EEG-BCI can be categorized based on the considered brain activity patterns~\cite{fazel2012p300}, i.e.: Event-Related Desynchronization/Synchronization; Steady State Visual Evoke Potentials (SSVEP); P300 component of the Event Related Potentials; Slow Cortical Potentials (SCPs).

In this work, similarly to~\cite{schroer2015autonomous}, we want to focus on the use of a EEG-BCI to control a robotic manipulator to perform operations as drinking or manipulating objects. However, here the BCI is operated on the base of the P300 paradigm %(and not on Event-Related Synchronization/Desynchronization)
that is a positive potential deflection on the ongoing brain activity at a latency of roughly 300 ms after the random occurrence of a desired target stimulus. This choice is motivated by the fact that P300-based BCIs %have recently been the focus of many studies,
are relatively easy to use for generating a control signal without complex training of the user, and have shown great potential to be used in several applications. From the robot motion control perspective, we do not base on path-planning algorithms %(as the Rapidly exploring Random Trees (RRTs) used in~\cite{schroer2015autonomous})
but we use the closed loop inverse kinematic approach presented in \cite{moe_etal_frontiers2016} that allows the coding of different kinds of high level actions required by the user and to manage of multiple tasks arranged in priorities and expressed as both set-based and equality-based constraints.
The effectiveness of the proposed architecture has been validated through experimental tests with a non-invasive BCI used to operate a 7 DOF lightweight robot manipulator. The video attached to the paper show the execution of a specific mission.

\section{Overall system architecture}  \label{sec:arch}
%Daniele (schema architettura, figura con schemi a blocchi)

The proposed system is composed of: a non-invasive BCI \emph{Emotiv Epoc+} that is a 14 channel wireless EEG-BCI; a 7 DOF lightweight robot manipulator \emph{Kinova Jaco\textsuperscript{2}}; a RGB-D sensor for human face and object recognition \emph{Microsoft Kinect One}. As schematically represented in Fig.~\ref{fig:scheme}, the different components are integrated through the Robot Operating System (ROS) framework where specific control, perception and communication modules have been developed for the application at hand.
%A sketch of the overall system architecture is shown in Fig.~\ref{fig:scheme}. 
The BCI is operated via a general-purpose software system BCI2000 that allows P300 experiments. The P300 operation paradigm allows the user to select via the BCI an option among a set. To each choice performed by the user is associated an action either to navigate the BCI2000 Graphical User Interface (GUI) or to send messages to external devices. In the latter case the BCI2000 opens an UDP/IP socket to send messages to a client process running on a linux machine that runs the ROS framework. The client process encodes the received message according to a beforehand established convention and send commands to the control module of the manipulator. The control module also collects messages from the perception module that allows to identify and localize objects or user's face in the workspace. In particular, Kinect's raw data are processed through OpenCV and PCL (Point Cloud Library) libraries. Finally, the control module of the manipulator implements a set-based closed loop inverse kinematics motion control algorithm that computes the corresponding desired joint velocities for the robotic manipulator.

\begin{psfrags}
\psfrag{d}[1pt][1pt]{$d$}
	\psfrag{LINUX_MACHINE}[][]{\footnotesize{LINUX MACHINE}}
	\psfrag{WINDOWS MACHINE}[][]{\footnotesize{WINDOWS MACHINE}}
	\psfrag{Planner}[][]{\footnotesize{Planner}}
	\psfrag{Socket UDP/IP}[][]{\footnotesize{Socket UDP/IP}}	
	\psfrag{UDP-ROS Bridge}[][]{\footnotesize{UDP-ROS Bridge}}	
	\psfrag{Kinematic Control}[][]{\footnotesize{Kinematic Control}}
	\psfrag{Raw Data}[][]{\tiny{Raw Data}}
	\psfrag{Workspace Tasks}[][]{\tiny{Workspace Tasks}}
	\psfrag{Desired Joint Velocities}[][]{\scriptsize{Desired Joint Velocities}}
	\psfrag{Control Signals}[][]{\scriptsize{Control Signals}}
	\psfrag{Brain Signals}[][]{\scriptsize{Brain Signals}}
	\psfrag{c/2}[][]{$d/2\,$}
\mypsfrag{5.5}{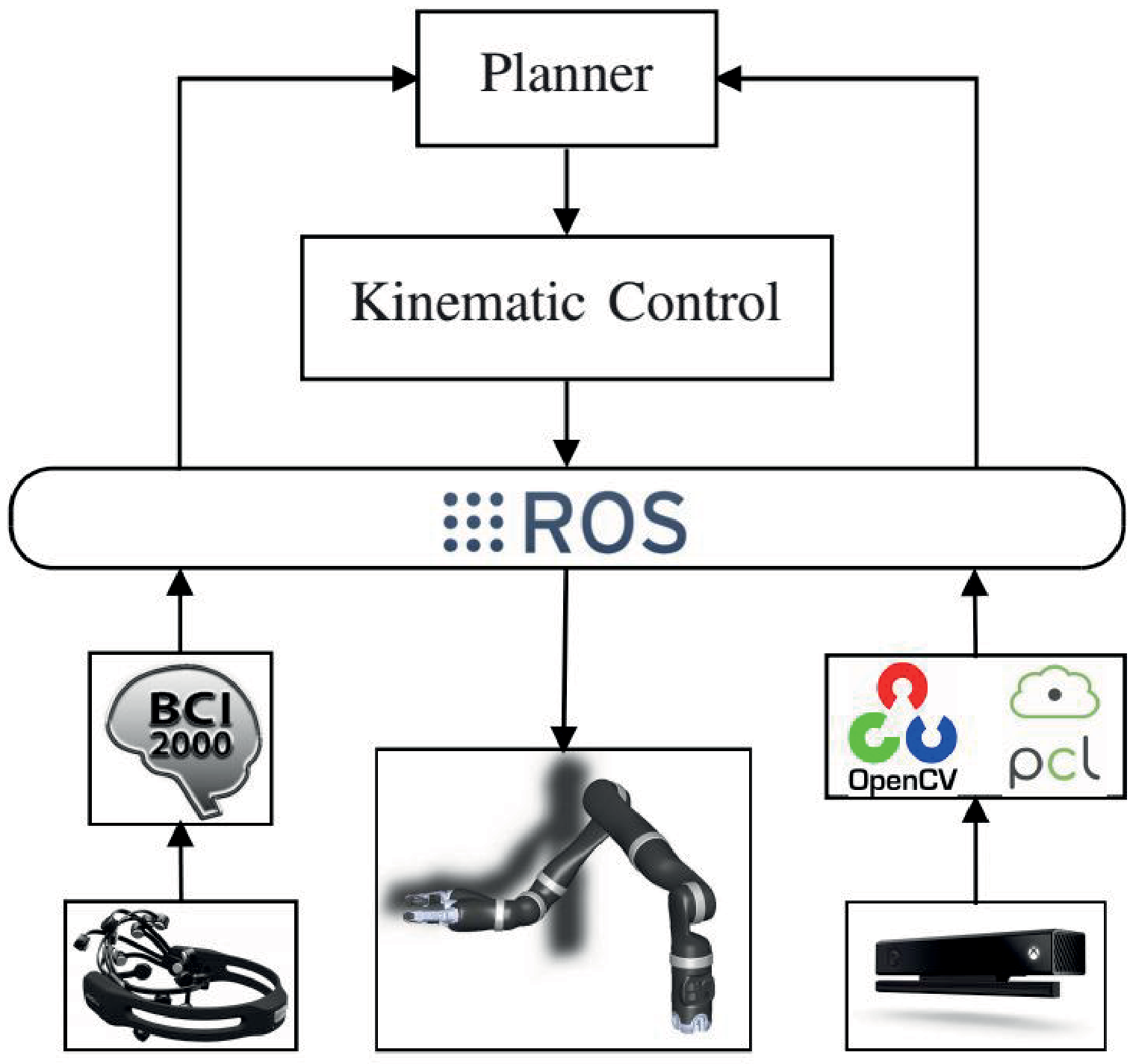}{-12pt}{Scheme of the overall architecture used in the experiment}{fig:scheme}
\end{psfrags}

\section{Operation of the EEG-BCI via P300 paradigm}\label{sec:bci}
%Filippo (Paradigma P300 e BCI2000)
The P300 potential is a component of the Event Related Potentials (ERPs), i.e. a fluctuation in the EEG generated by the electrophysiological response to a significant sensorial stimulus or event. In particular, the P300 is the largest ERP component, and it consists of a positive shift in the EEG signal approximately 300-400ms after a task relevant stimulus. The P300 potential can be generated during an oddball paradigm by which the user is subject to a sequence of events (e.g. visual stimulus) that can be categorized into two classes, one less frequent than the other. The frequent events are named standard stimulus, while the unfrequent events are named target stimulus. When the user distinguishes a target stimulus from a standard one, this generates the P300 peak about 300 ms after stimulus onset.

The P300 potential has been used as the basis for a BCI system in many studies. The classical format presents the user with a matrix of characters whose rows and columns flash successively and randomly at a rapid rate. The user selects a character by focusing attention on it and counting how many times it flashes. The row or column that contains this character evokes a P300 response, whereas all others do not. After a proper training, the computer can determine the desired row and column with the highest P300 amplitude, and thus the desired character.

The BCI considered in the proposed architecture is operated via a general-purpose software system named BCI2000 that, among the different features, allows BCI data acquisition, signal processing, stimulus presentation, and P300 experiments. In particular, for the application at hand, we want to allow an user to operate a robotic manipulator to achieve manipulation actions on some of the objects present in its workspace.% (e.g. move an object from one position to another or take a bottle to let the user drink).
Thus, we rely on BCI2000 P300 functionalities to allow the user to generate commands through a P300 based Graphical User Interface (GUI). This GUI has been realized structuring the array of flashing elements in a multi-layered structure; i.e. selecting an element from a starting array, the GUI can both switch to a different layer (i.e. with a different array of elements) and send commands to the manipulator through
%. Indeed, it is possible to associate to the different elements
callback functions that open UDP/IP socket to communicate with a remote machine.

Fig.~\ref{fig:interfaceBCI} shows an example of the structure of the developed GUI where the user can select operations as: select an object, pause/sleep the P300 GUI, move to the home level, command an action to the robot, stop the robot, etc. It can be noticed that characters, commonly used for the P300 paradigm, have been replaced by flashing icons more intuitive for the user.

\begin{psfrags}
\psfrag{d}[1pt][1pt]{$d$}
	\psfrag{Obj1}[][]{\footnotesize{Obj1}}
	\psfrag{Obj2}[][]{\footnotesize{Obj2}}
	\psfrag{Sleep}[][]{\footnotesize{Sleep}}
	\psfrag{Home}[][]{\footnotesize{Home}}	
	\psfrag{Action1}[][]{\footnotesize{Action1}}	
	\psfrag{Action2}[][]{\footnotesize{Action2}}
	\psfrag{SubAction1}[][]{\tiny{SubAction1}}
	\psfrag{SubAction2}[][]{\tiny{SubAction2}}
	\psfrag{Emergency Stop}[][]{\scriptsize{Emergency Stop}}
	\psfrag{Emergency Stop}[][]{\scriptsize{E}}
	\psfrag{Back}[][]{\scriptsize{Back}}
	\psfrag{Object Selection}[][]{\tiny{\bf{Object Selection}}}
	\psfrag{Action Selection}[][]{\tiny{\bf{Action Selection}}}
\mypsfrag{5.5}{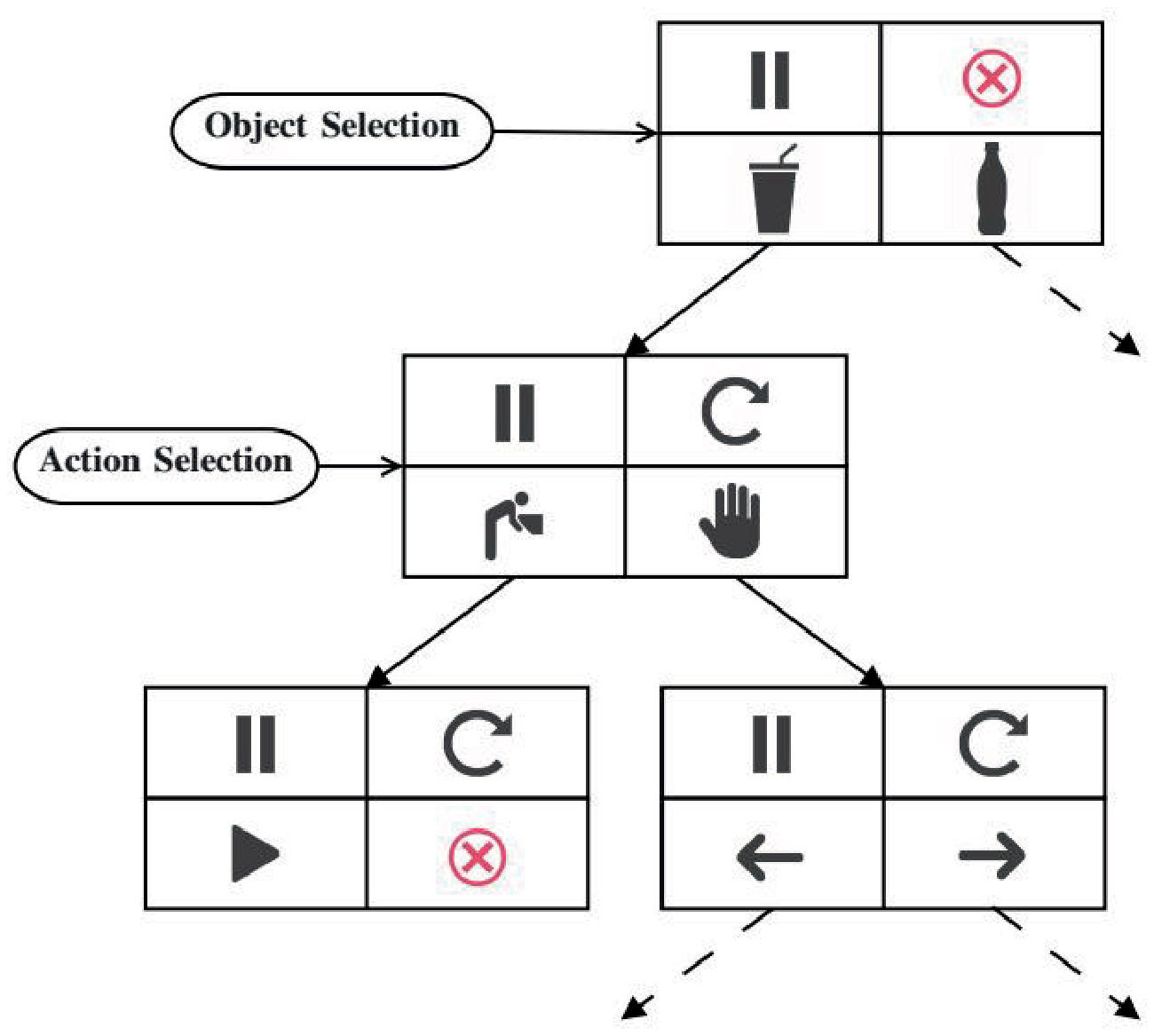}{-12pt}{Scheme of the developed BCI2000 user interface.}{fig:interfaceBCI}
\end{psfrags}

Before using the developed GUI, the user should train the P300 software following a specific procedure that consists in selecting via oddball paradigm a set of characters in a predefined order. The BCI2000 software uses a genetic algorithm for the training of a Stepwise Linear Discriminant Analysis (SWLDA) binary classifier used in the operation mode, and it generates a specific profile file for the user. %Then, in the operation mode, BCI2000 software uses a Stepwise Linear Discriminant Analysis (SWLDA) binary classifier to perform the P300 experiment.

The used BCI  is an Epoc+ produced by Emotiv, that is a low-cost non-invasive BCI offering high resolution (14 bits, 1 LSB = 0.51$\mu$V) multi-channels signals, and that provides access to dense array, high quality, raw EEG data. The 14 electrodes of the neuroheadset (that generate signals with a bandwidth of 0.2~–-~43Hz) are located according to the 10-20 international system. %The neuroheadset also contains accelerometers and gyroscopes that are not used for the purpose of this work.

\section{Motion control of the robotic manipulator}\label{sec:manip}
%Paolo  (controllo cinematico set based)

The robot manipulator receives as input from the BCI high-level commands containing information about the selected object and the kind of action to perform. The set of possible actions is predefined but their execution depend on the information collected on-line about the workspace, e.g. the placement of objects, the position of the user's face, the possible presence of obstacles.

Each action is coded via a set of elementary tasks, each described by a suitable task function of the system state, and arranged in priority order as described in \cite{AntArrChi_ISR08}. The reference system velocity is computed inverting the task function at a kinematic level and by projecting the contribution of each task into the null space of the higher priority ones so as to remove the velocity components that would conflict with it.

For a general robotic system with $n$ Degrees of Freedom, the state is described by the joint values $\bfq = \left[q1,q2,\dots,qn\right]^T\in \mathbb{R}^{\it{n}}$.  Let us consider a generic $m$-dimensional task function $\bf{\boldsymbol{\sigma}(q)} \in \mathbb{R}^{\it{m}}$. The following differential relationship holds:
\begin{center}
$\bf{\dot{\boldsymbol{\sigma}}(q)}=\bf{J(q)}\bf{\dot{q}}\quad,$
\end{center}
where $\bf{J(q)} \in \mathbb{R}^{\it{m\times n}}$ is the task Jacobian matrix, and $\bf{\dot{q}}$ is the system velocity. The reference velocity that brings the task value $\bf{\boldsymbol{\sigma}}$ to a desired $\bf{\boldsymbol{\sigma}_d}$ can be computed as:
\begin{equation}\label{eq:diff}
\bf{\dot{q}}=\bf{J}^{\dagger}(\bf{\dot{\boldsymbol{\sigma}}_d}+\bf{K}\bf{\tilde{\boldsymbol{\sigma}}}),
\end{equation}
where $\bf{K}$ is a positive-definite matrix of gains, and ${\tilde{\bfsigma}}=\bfsigma_d-\bfsigma$ is the task error.
If the system is redundant with respect to the task dimension ($n>m$) it is possible to fulfill multiple tasks simultaneously.
Defining a priority among the $h$ tasks composing an action, the reference system velocity can be computed as:
\begin{equation}\label{eq:nsb}
\bf{\dot{q}}=\bf{\dot{q}_1}+\bf{N_1}\bf{\dot{q}_2}+\dots+\bf{N_{1,h-1}}\bf{\dot{q}_h}\quad,
\end{equation}
where $\bf{\dot{q}_i}$ is the reference velocity that fulfills the $i$-th task and $\bf{N_{1,i}}$ is the null space of the augmented Jacobian:

\begin{equation} \label{eq:jaco}
\bf{J_{1,i}}=
\begin{bmatrix}
\bf{J_1}^T &\bf{J_2}^T &\dots &\bf{J_i}^T
\end{bmatrix}^T.
\end{equation}

This framework has  been recently extended in \cite{moe_etal_frontiers2016,Moe_cdc2015} in order to handle also set-based tasks, 
i.e. monodimensional tasks requiring their value to lie in a set of values  $\boldsymbol{\bf\mathscr{D}}$ rather than assuming a specific one. %, i.e. below or above a certain threshold.
Classic set-based tasks for a robotic manipulator are mechanical joints limits, obstacle avoidance, and arm manipulability. The considered method allows to simultaneously control a hierarchy composed of both equality-based and set-based tasks. In particular, while the equality-based tasks are always active, the set-based tasks can be activated or deactivated depending on the operational conditions. For each set-based task it is indeed necessary to set different reference values: physical thresholds $\sigma_{\text{max}}$ ($\sigma_{\text{min}}$), activation thresholds $\sigma_{\text{max}}-\epsilon$ ($\sigma_{\text{min}}+\epsilon$), and safety values $\sigma_{s,u}$ with $\sigma_{\text{max}}-\epsilon<\sigma_{s,u}<\sigma_{\text{max}}$ ($\sigma_{s,l}$ with $\sigma_{\text{min}}<\sigma_{s,l}<\sigma_{\text{min}}+\epsilon$ ).
With reference to Figure \ref{fig:thre}, as long as the threshold for a specific set-based task is satisfied (i.e. $\sigma_{min}+\epsilon<\sigma<\sigma_{max}-\epsilon$), the task is removed from the hierarchy and the solution that fulfills the other tasks is computed. When its threshold is violated (i.e. $\sigma<\sigma_{\text{min}}+\epsilon \quad || \quad\sigma>\sigma_{\text{max}}-\epsilon$ ), the task is re-inserted in the hierarchy and it is transformed into an equality-based task. The desired value of the task function is set as:
 \begin{equation}
\sigma_d=\left\{\begin{array}{l l}
\sigma_{s,u} &\text{if $\sigma_{max}-\epsilon<\sigma<\sigma_{max}$}\\
\sigma_{s,u} &\text{if $\sigma_{min}<\sigma<\sigma_{min}+\epsilon$}
%\sigma_{\text{max}}-\epsilon & \text{if $\sigma_{\text{max}}-\epsilon<\sigma<\sigma_{\text{max}}$}
 \end{array} \right. %$, if the upper bound is violated
\end{equation}

It is important that $\sigma_{\text{max}}-\epsilon<\sigma_{s,u}<\sigma_{\text{max}}$ ($\sigma_{\text{min}}<\sigma_{s,l}<\sigma_{\text{min}}+\epsilon$), in order to avoid undesirable system behaviors as chattering due to intermittent activation/deactivation of tasks caused by quantization errors or
sensor noise in the task value computation.

\begin{psfrags}
\mypsfrag{8}{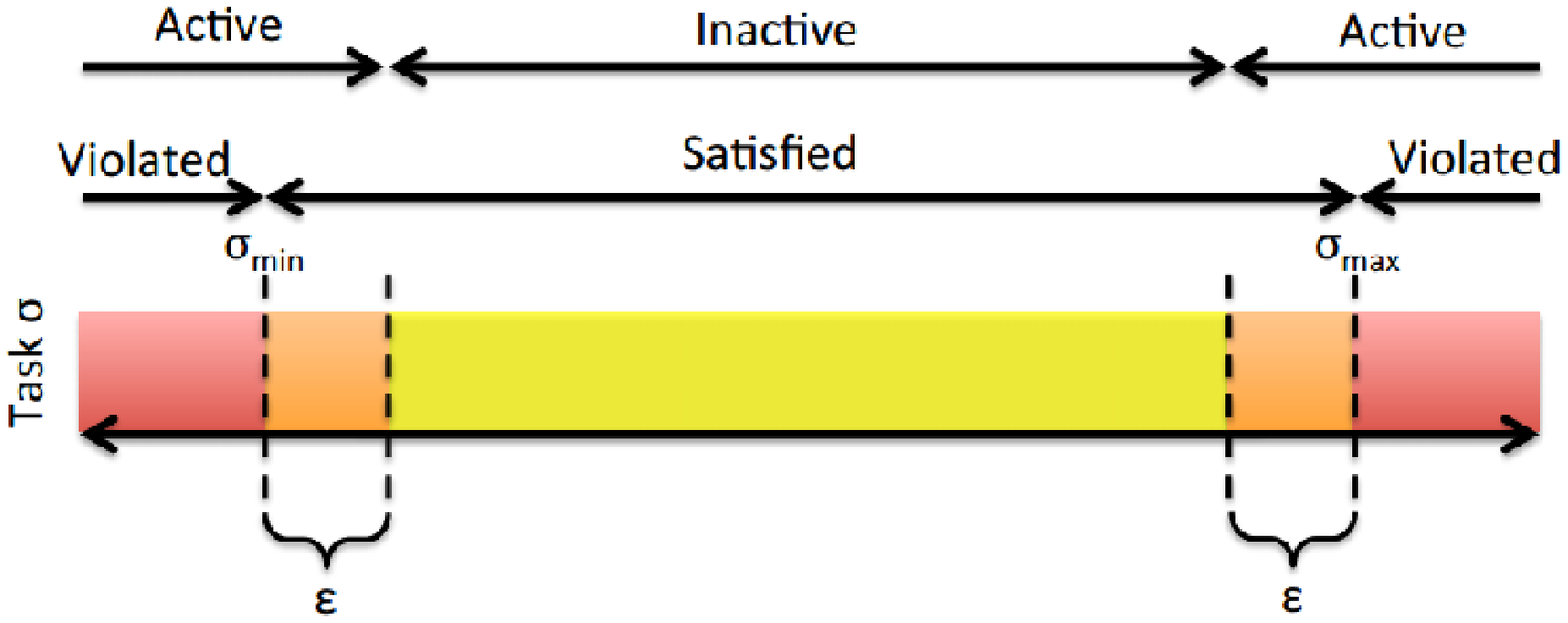}{-12pt}{Activation and physical thresholds of a set-based task}{fig:thre}
\end{psfrags}

It is not always necessary to compute the solution of the hierarchy containing all the active set-based tasks, because
the desired system velocity computed applying (\ref{eq:nsb}) to a hierarchy that do not contain a specific set-based task
could bring such task anyway to its valid set of values. In that case, the set-based task can be removed from the active task hierarchy on the base of the following algorithm.

\subsection{Set-based activation/deactivation algorithm }
The algorithm can be divided into four main steps:
\begin{enumerate}
\item Create the active task hierarchy
\item Compute the solutions
\item Compute projections of the solutions
\item Choice of the solution
\end{enumerate}

Starting from a  hierarchy  $\boldsymbol{\bf\mathscr{H}}$ of mixed set-based and equality-based tasks, in the first step the hierarchy  $\boldsymbol{\mathscr{A}}$ containing all the active tasks is created, that means $\boldsymbol{\mathscr{A}}$  is composed of all the equality-based tasks and all the set-based tasks that exceed the activation thresholds.

Given that we can not know a-priori which solution would make all the set-based task to stay in their specific set of values,
it is necessary to build a solutions tree, by computing all the solutions given by (\ref{eq:nsb}) on all the possible task hierarchies obtained by inserting and
removing all the set-based active tasks, i.e for an active hierarchy composed by $n_a$ set-based tasks, it is necessary
to compute $2^{n_a}$ solutions and store them in a set $\boldsymbol{\bf\mathscr{S}}$. 

Then we have to select and store in a set $\boldsymbol{\bf\mathscr{P}}$ among all the solutions in $\boldsymbol{\bf\mathscr{S}}$, the ones that satisfy all the active set-based tasks while fulfilling also all the equality-based tasks. It is
possible to check it by projecting each solution in $\boldsymbol{\bf\mathscr{S}}$ into all the active set-based task spaces: a solution $\bf\dot{q}_j$ fulfills a set-based task $\sigma_k$ if $\bf{J_k}\dot{\bf{q}}_j<0$ ($\bf{J_k}\dot{\bf{q}}_j>0$). 

Finally, one solution among the ones in $\boldsymbol{\bf\mathscr{P}}$ has to be chosen as the desired system velocity $\bf{\dot{q}_d}$. It is important to notice that the set
$\boldsymbol{\bf\mathscr{P}}$ is never empty, because there will always be the solution that takes into account all the set-based tasks. If $\boldsymbol{\bf\mathscr{P}}$ contains more than one solution the highest-norm one is chosen, being the less conservative in
terms of system velocity.

\section{Robotic perception software}\label{sec:perc}
%Paolo  (percezione oggetti/marker e volto)
The robotic platform has to be capable to interact with the environment and with
the user, thus a perception system is needed to make the robot aware of the position of the
user and of the possible objects. Objects in the environment are labelled with markers, and their detection and tracking are performed by resorting to the ArUco library \cite{Aruco2014}, a well-known OpenCV module specifically designed for this kind of operations. Objects positions are computed in Kinect reference frame and then transformed into the manipulator reference frame by mean of a transformation matrix computed with a preliminary calibration. % This operation needs a preliminary environment calibration, i.e. the computation of the transformation matrix $T_{kin}^{rob}$ between the Kinect frame an the robot one.

For the case study presented in the following section we are interested in detecting and tracking also the mouth of the BCI's user. Such operation has been divided in two steps: the first one consists in detecting the mouth position in the 2D plane of the RGB image taken from the Kinect sensor, and then in estimating the distance from the point cloud.
First of all the image in Full HD resolution is acquired from the sensor, and then it is
downsampled in order to reduce the computational load and to make the algorithm more suitable for a real-time application. Then a Viola-Jones algorithm is applied to recognize the face in the scene, specifically using the Haar features
\cite{castrillon2011comparison}, \cite{mistry2013literature}. Among all the faces detected
within the image, only the closest one is selected. The face image is then split
into two parts, and only the lower one is taken into account for the following
computations. The Haar features are once again applied in order to find the coordinates of
the mouth in the image frame.

The second step is the computation of the 3D coordinates of the center of the mouth. The Full HD Point Cloud is acquired from the Kinect, and then filtered by a Voxel Grid filter in order to reduce the number of points to be computed. The points belonging to
the selected area of the image are extracted and then a mean for the $x$, $y$ and $z$ coordinates of the center of the mouth is computed. %In Fig.~\ref{fig:face}.b there's a screenshot of the detected area of the mouth in the RGB image superimposed on the depth cloud.

\section{Experimental case study}\label{sect:sim}
%Daniele/Paolo  (schema interfaccia bci, azioni implementate, risultati)
In the considered experimental case study, we want to allow a user to command the robot through the BCI to move objects in preselected positions or to bring a bottle to his mouth. To the purpose, a specific GUI has been developed in the BCI2000 framework according to the scheme in Fig.~\ref{fig:interfaceBCI} and that result navigable via P300 BCI paradigm. %The different layers are reported in  Fig.~\ref{fig:interfaceBCI} a GUI  BCI user interface according to scheme in Fig.~\ref{fig:scheme} has been implemented. More in detail, the interface
The GUI composed of the four different layers shown in Fig.~\ref{fig:bci}. Referring to this figure, the first selection (image~1) represents the object selection layer, from where the user can choose the object to manipulate, in this case a bottle coke and one of water. Furthermore there is the possibility to pause the interface program execution (represented by the ``pause'' icon) and to send a stop signal to the manipulator in case of emergency (represented by the red cross). The second selection (image~2) represents the action selection layer, in fact after the object choice the user can decide which action to perform, i.e. whether to drink or to move the chosen bottle. Even in this case there is the possibility to pause and to return to the bottle choice (represented by the round arrow). If the user decides to drink, the interface switches directly to the fourth selection (image~4) and pauses itself. In this phase the user can decide to resume the interface program (represented by the ``play'' icon), to send a stop signal to the robot manipulator, to return to the bottle choice or to go back to the previous selection(represented by the ``-1'' icon). Instead, if the user decides to take and move the bottle, the interface switches on the third selection (image~3) that represents the sub-action selection. In this phase the user can decide to move the bottle on the left or on the right in preassigned positions. After the choice of the location, the interface advances to the fourth selection. Once object and actions are selected, the GUI sent a message to the robot control with the details of the choice performed by the user.  
\begin{psfrags}
\psfrag{d}[1pt][1pt]{$d$}
	\psfrag{LINUX_MACHINE}[][]{\footnotesize{LINUX MACHINE}}
	\psfrag{WINDOWS MACHINE}[][]{\footnotesize{WINDOWS MACHINE}}
	\psfrag{Planner}[][]{\footnotesize{Planner}}
	\psfrag{Socket UDP/IP}[][]{\footnotesize{Socket UDP/IP}}	
	\psfrag{UDP-ROS Bridge}[][]{\footnotesize{UDP-ROS Bridge}}	
	\psfrag{Kinematic Control}[][]{\footnotesize{Kinematic Control}}
	\psfrag{Raw Data}[][]{\tiny{Raw Data}}
	\psfrag{Workspace Tasks}[][]{\tiny{Workspace Tasks}}
	\psfrag{Desired Joint Velocities}[][]{\scriptsize{Desired Joint Velocities}}
	\psfrag{Control Signals}[][]{\scriptsize{Control Signals}}
	\psfrag{Brain Signals}[][]{\scriptsize{Brain Signals}}
	\psfrag{c/2}[][]{$d/2\,$}
\mypsfrag{7.5}{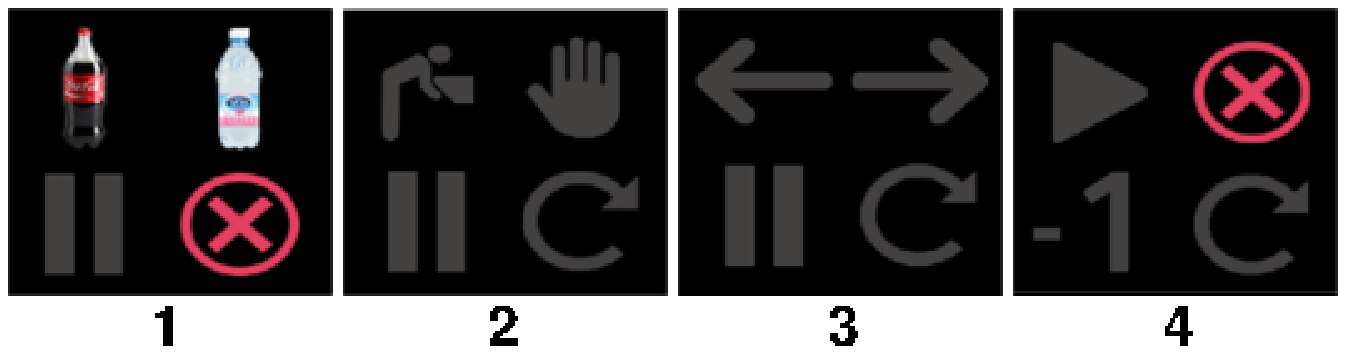}{-12pt}{Different choices of the BCI user interface}{fig:bci}
\end{psfrags}

In the following there is the description of the results for the two kinds of performed operations.
\subsection{``Move'' operation}
For the first experiment the user has been asked to choose to move the water bottle on the right. The high-level command built by the
BCI user interface is sent to the manipulator that  autonomously fulfills the operation. The chosen task hierarchy for the operation is:
\begin{enumerate}
\item {\bf Fourth joint mechanical limit:} a maximum limit of 5.5\ts rad and a minimum limit of 0.7\ts rad have been set for the fourth joint of the manipulator in order to avoid that it hits its own structure
\item {\bf Obstacle avoidance:} as the operator chooses a bottle, the other automatically becomes an obstacle that the end-effector of the manipulator needs to avoid. A minimum threshold on the distance between the end-effector and the obstacle of 25\ts cm has been set.
\item {\bf End-effector position and orientation:} a sequence of target waypoints for the end-effector position and orientation has been chosen in order to make the manipulator effectively grasp the selected bottle and to move it in a specific position.

\end{enumerate}

\begin{psfrags}
\mypsfragtwocolumn{17.5}{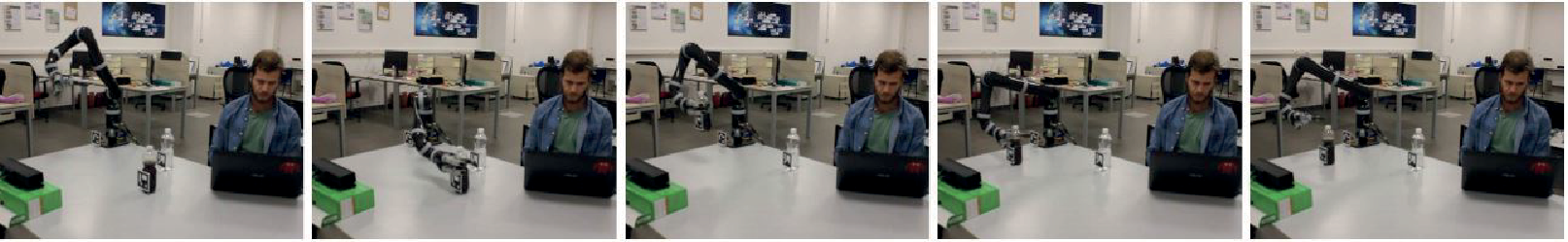}{-12pt}{Frames of the ``Move'' operation execution}{fig:frameTake}
\end{psfrags}

Fig. \ref{fig:move01} shows the end-effector position and orientation error during the entire operation. Fig. \ref{fig:move02} shows the set-based tasks values, and it can be seen that all the thresholds for all the set-based tasks are respected. Fig \ref{fig:frameTake} shows a sequence of snapshots of a performed moving mission.

\begin{psfrags}
\psfrag{2000}[][]{\scriptsize{20}}
\psfrag{4000}[][]{\scriptsize{40}}
\psfrag{6000}[][]{\scriptsize{60}}
\psfrag{8000}[][]{\scriptsize{80}}
\psfrag{10000}[][]{\scriptsize{100}}
\psfrag{12000}[][]{\scriptsize{120}}
\psfrag{0}[][]{\scriptsize{0}}
\psfrag{1}[][]{\scriptsize{1}}
\psfrag{-1}[][]{\scriptsize{-1}}
\psfrag{-2}[][]{\scriptsize{-2}}
\psfrag{0.5}[][]{\scriptsize{0.5}}
\psfrag{-0.5}[][]{\scriptsize{-0.5}}
\psfrag{[s]}[][]{\scriptsize{\text{[s]}}}
\psfrag{[m]}[][]{\scriptsize{\text{[m]}}}
\psfrag{a1}[][]{\scriptsize{\text{a)}}}
\psfrag{b1}[][]{\scriptsize{\text{b)}}}
\mypsfrag{8}{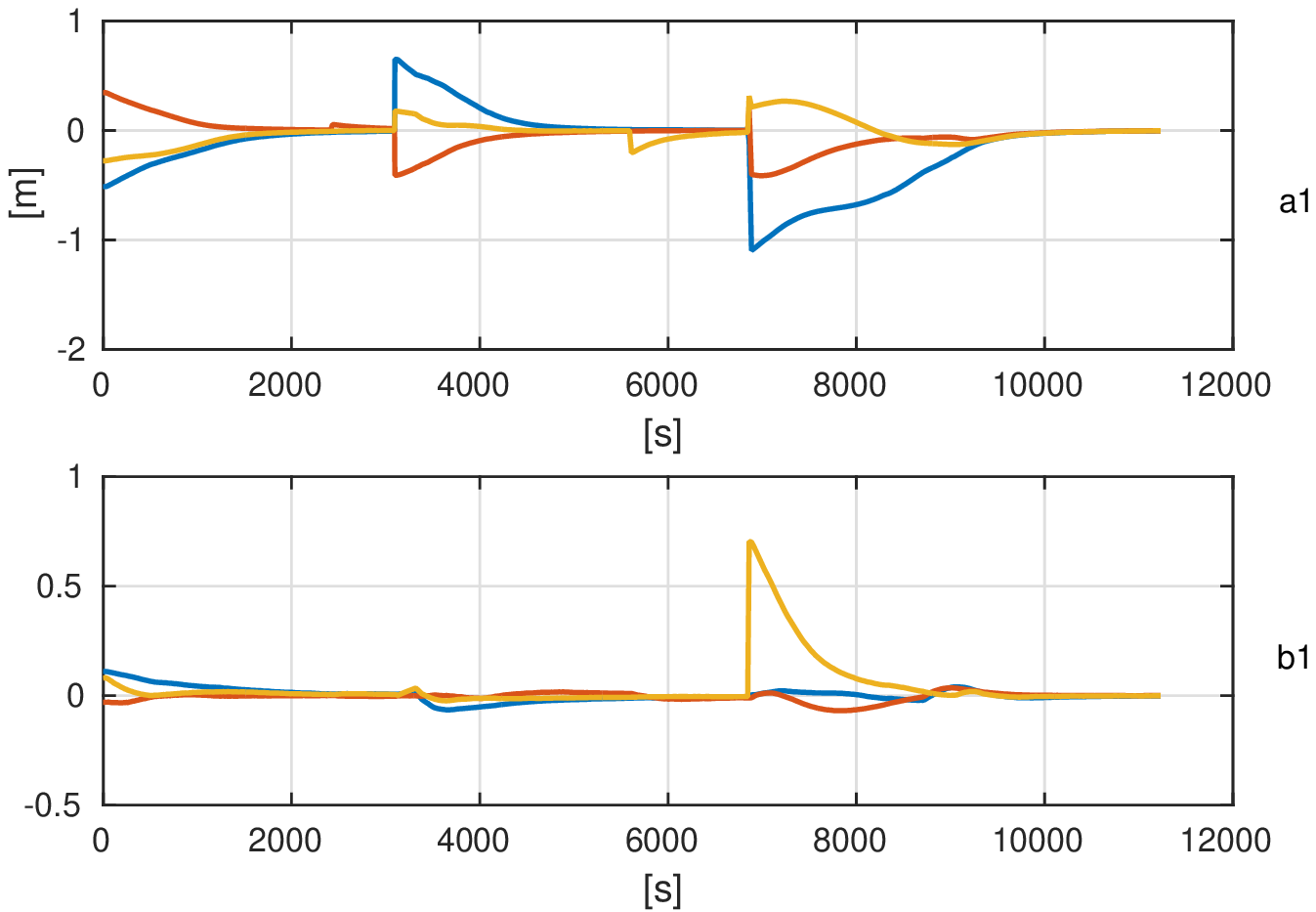}{-12pt}{``Move'' operation: a) End-effector position error on $x$-axis (blue), $y$-axis (yellow) and $z$-axis (red) during the operation; b) End effector orientation error on $x$-axis (blue), $y$-axis (yellow) and $z$-axis (red) during the operation}{fig:move01}
\end{psfrags}

\begin{psfrags}
\psfrag{2000}[][]{\scriptsize{20}}
\psfrag{4000}[][]{\scriptsize{40}}
\psfrag{6000}[][]{\scriptsize{60}}
\psfrag{8000}[][]{\scriptsize{80}}
\psfrag{10000}[][]{\scriptsize{100}}
\psfrag{12000}[][]{\scriptsize{120}}
\psfrag{0}[][]{\scriptsize{0}}
\psfrag{0.5}[][]{\scriptsize{0.5}}
\psfrag{1}[][]{\scriptsize{1}}
\psfrag{1.5}[][]{\scriptsize{1.5}}
\psfrag{2}[][]{\scriptsize{2}}
\psfrag{4}[][]{\scriptsize{4}}
\psfrag{6}[][]{\scriptsize{6}}
\psfrag{[rad]}[][]{\scriptsize{\text{[rad]}}}
\psfrag{[s]}[][]{\scriptsize{\text{[s]}}}
\psfrag{[m]}[][]{\scriptsize{\text{[m]}}}
\psfrag{a1}[][]{\scriptsize{\text{a)}}}
\psfrag{b1}[][]{\scriptsize{\text{b)}}}
\psfrag{c1}[][]{\scriptsize{\text{c)}}}
\mypsfrag{8}{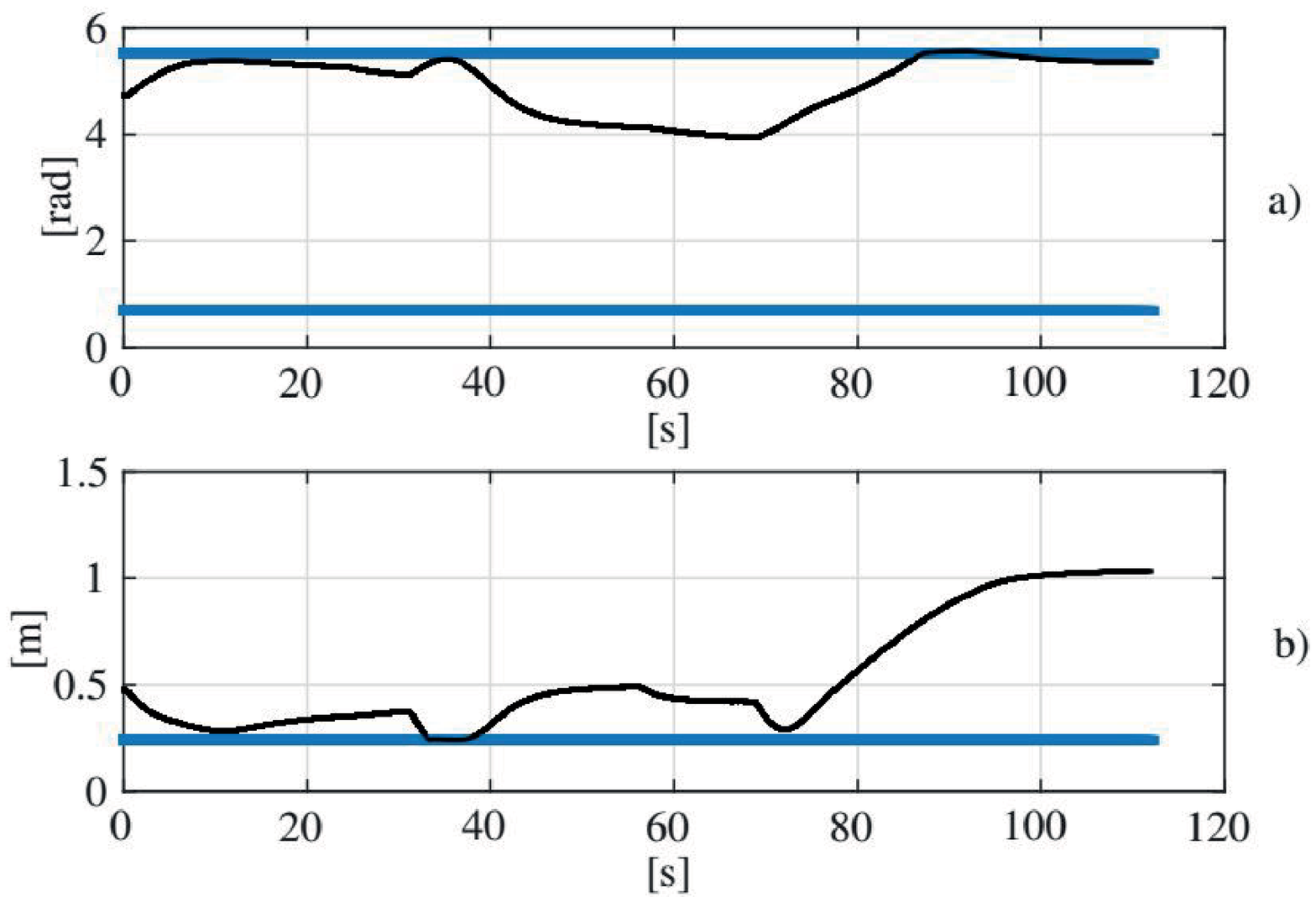}{-12pt}{``Move'' operation: a) Fourth joint position (black) and limits (blue) over time; b) Distance from the obstacle (blue) and minimum distance (black) over time}{fig:move02}
\end{psfrags}

\subsection{``Drink'' operation}

For the second experiment the user has been asked to choose to drink from the water bottle. The task hierarchy is:
% slightly different with respect to the one chosen for the ``move'' operation:
\begin{enumerate}
\item {\bf Fourth joint limit:} same as for the first experiment
\item {\bf Second joint limit:} a maximum limit of 5.1 rad and a minimum limit of 1.9 rad has been chosen for the second joint position, in order to avoid a collision between the ``elbow'' of the manipulator and the table on which the objects are placed.
\item {\bf Obstacle avoidance:} same as the first experiment
\item {\bf Bottle top position and orientation:} in this case it is necessary to control the position and the orientation of the cap of the grasped bottle rather than the end-effector ones. Similarly to the first experiment, a proper sequence of target waypoints and orientations have been assigned in order to fulfill the operation: grasp the bottle, get it close to the operator's mouth, make him drink, and reposition the bottle on the table.
\end{enumerate}

\begin{psfrags}
\mypsfragtwocolumn{17.5}{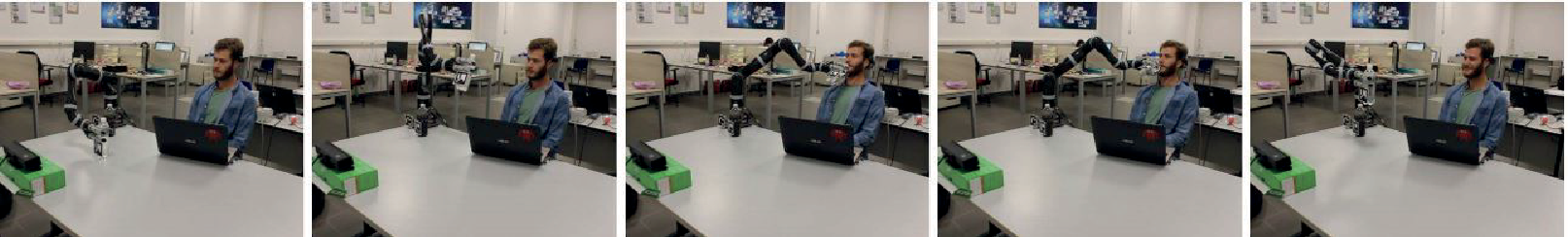}{-12pt}{Frames of the ``Drink'' operation execution}{fig:frameDrink}
\end{psfrags}
Fig. \ref{fig:drink01} and Fig. \ref{fig:drink02} show the results of the experiment. The bottle cap follows the desired positions and orientations with small errors, while respecting all the thresholds for all the set-based tasks,  effectively fullfilling the operation. Fig \ref{fig:frameDrink} shows a sequence of snapshots of a performed drinking mission.

\begin{psfrags}
\psfrag{2000}[][]{\scriptsize{20}}
\psfrag{4000}[][]{\scriptsize{40}}
\psfrag{6000}[][]{\scriptsize{60}}
\psfrag{8000}[][]{\scriptsize{80}}
\psfrag{10000}[][]{\scriptsize{100}}
\psfrag{12000}[][]{\scriptsize{120}}
\psfrag{14000}[][]{\scriptsize{140}}
\psfrag{0}[][]{\scriptsize{0}}
\psfrag{2}[][]{\scriptsize{2}}
\psfrag{4}[][]{\scriptsize{4}}
\psfrag{6}[][]{\scriptsize{6}}
\psfrag{0.5}[][]{\scriptsize{0.5}}
\psfrag{-0.5}[][]{\scriptsize{-0.5}}
\psfrag{1}[][]{\scriptsize{1}}
\psfrag{[s]}[][]{\scriptsize{\text{[s]}}}
\psfrag{[m]}[][]{\scriptsize{\text{[m]}}}
\psfrag{a1}[][]{\scriptsize{\text{a)}}}
\psfrag{b1}[][]{\scriptsize{\text{b)}}}
\psfrag{c1}[][]{\scriptsize{\text{c)}}}

\mypsfrag{8}{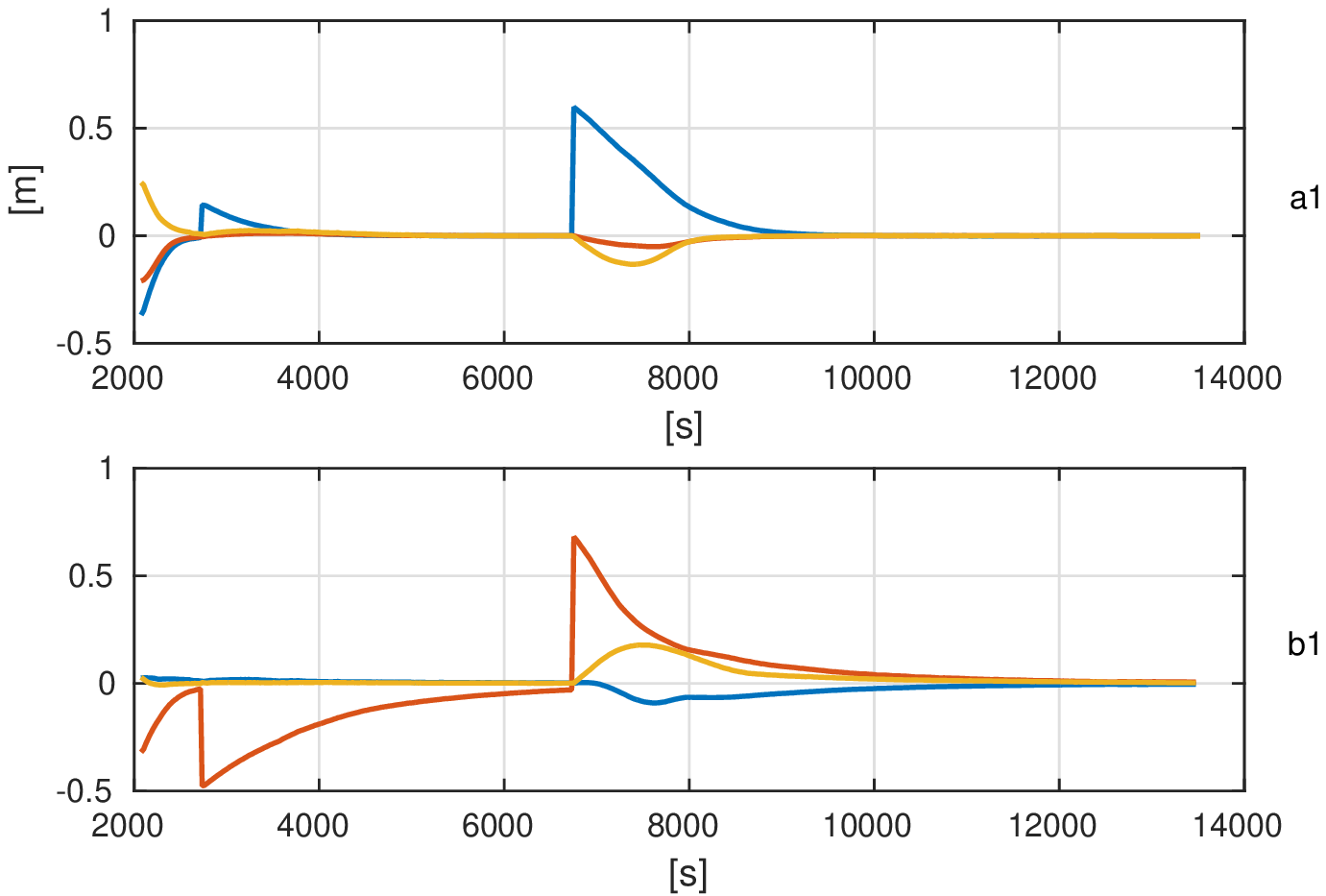}{-12pt}{``Drink'' operation: a) Bottle top position error on $x$-axis (blue), $y$-axis (yellow) and $z$-axis (red) during the operation; b) Bottle top orientation error on $x$-axis (blue), $y$-axis (yellow) and $z$-axis (red) during the operation}{fig:drink01}
\end{psfrags}

\begin{psfrags}
\psfrag{2000}[][]{\scriptsize{20}}
\psfrag{4000}[][]{\scriptsize{40}}
\psfrag{6000}[][]{\scriptsize{60}}
\psfrag{8000}[][]{\scriptsize{80}}
\psfrag{10000}[][]{\scriptsize{100}}
\psfrag{12000}[][]{\scriptsize{120}}
\psfrag{14000}[][]{\scriptsize{140}}
\psfrag{0}[][]{\scriptsize{0}}
\psfrag{1}[][]{\scriptsize{1}}
\psfrag{-1}[][]{\scriptsize{-1}}
\psfrag{2}[][]{\scriptsize{2}}
\psfrag{4}[][]{\scriptsize{4}}
\psfrag{6}[][]{\scriptsize{6}}
\psfrag{0.5}[][]{\scriptsize{0.5}}
\psfrag{-0.5}[][]{\scriptsize{-0.5}}
\psfrag{[s]}[][]{\scriptsize{\text{[s]}}}
\psfrag{[m]}[][]{\scriptsize{\text{[m]}}}
\psfrag{[rad]}[][]{\scriptsize{\text{[rad]}}}
\psfrag{a1}[][]{\scriptsize{\text{a)}}}
\psfrag{b1}[][]{\scriptsize{\text{b)}}}
\psfrag{c1}[][]{\scriptsize{\text{c)}}}
\mypsfrag{8}{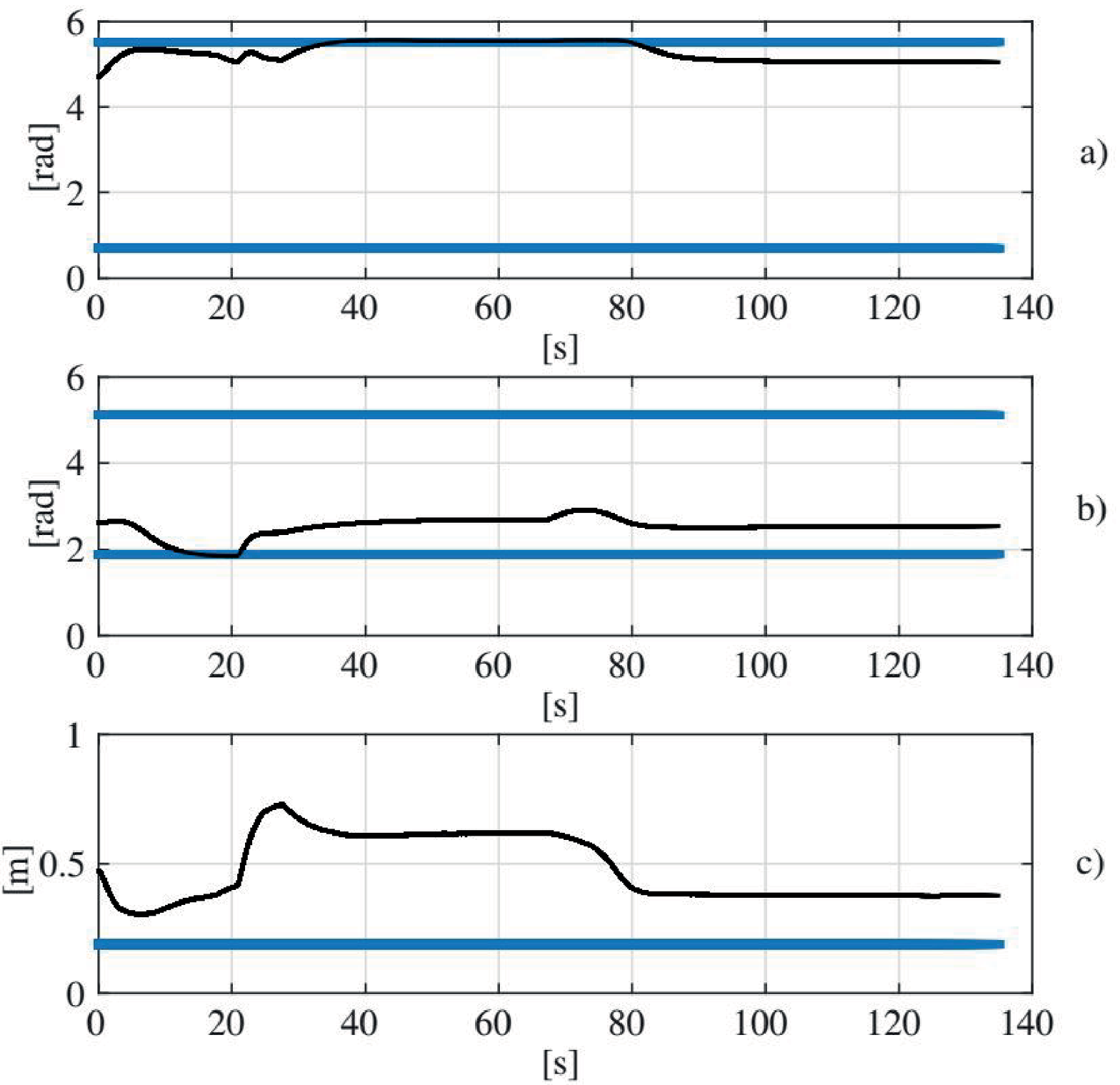}{-12pt}{``Drink'' operation: a) Fourth joint position (black) and limits (blue) over time;  b) Second joint position (black) and limits (blue) over time;  c)  Distance from the obstacle (blue) and minimum distance (black) over time}{fig:drink02}
\end{psfrags}

\section{Conclusions}\label{sect:conc}
This paper shows an architecture for an assistive robotic system aimed at helping users with sever motion disabilities in daily-life operations. The proposed system relies on a BCI based on P300 paradigm for the high level command detection, a Kinect One sensor for the environment perception and a Kinova Jaco\textsuperscript{2} lightweight robot manipulator for performing the manipulation tasks. Details of the software modules and of the specific motion control algorithm applied for the application at hand have been described, and an experimental case study involving two different kinds of operations have been reported to prove the effectiveness of the developed system.

Further efforts will mainly concern the improvement of the perception system, the user interface and the motion/interaction control of the robot. More in detail, we want to substitute the marker-based object detection algorithm with a detection algorithm based on the object geometrical shapes. Moreover, we want to make the BCI GUI capable of dynamically changing the structure of the layers to adapt to the environment scene and to replace object icons with images dynamically taken from the Kinect. For the robot control, future activity will focus on more sophisticated obstacle avoidance algorithm to add an user's safety levels and we will consider more suitable control algorithm for the interaction with the environment and for collisions detection.

%---------------------------------------------------------------------------------------

\section*{Acknowledgments}
The Authors want to thank Alessandro Bria, Gianfranco Miele, Mario Fresilli and Gianluca Mangiapelo for the provided support. %  in the development and testing of the differnt .
%The research leading to these results has received funding from the Italian
%Government, under Grant FIRB - Futuro in ricerca 2008 n.~RBFR08QWUV  (project
%NECTAR), PRIN 2009
%n.~20094WTJ29 (project RoCoCo).

\bibliography{biblio_bci}

%\appendix

\end{document}